# DOMAIN-TRANSFERABLE METHOD FOR NAMED ENTITY RECOGNITION TASK

Vladislav Mikhailov[1, 2] and Tatiana Shavrina[1,2]

[1]Sberbank, Moscow, Russia
[2]Higher School of Economics, Moscow, Russia

## ABSTRACT

*Named Entity Recognition (NER) is a fundamental task in the fields of natural language processing and information extraction. NER has been widely used as a standalone tool or an essential component in a variety of applications such as question answering, dialogue assistants and knowledge graphs development. However, training reliable NER models requires a large amount of labelled data which is expensive to obtain, particularly in specialized domains. This paper describes a method to learn a domain-specific NER model for an arbitrary set of named entities when domain-specific supervision is not available. We assume that the supervision can be obtained with no human effort, and neural models can learn from each other. The code, data and models are publicly available.*

## 1. INTRODUCTION

Named Entity Recognition (NER) is a rapidly developing NLP task that aims to extract mentions of entities from texts and label them with the predefined semantic types such as PER (person), LOC (geographical location), ORG (organization), etc. Regarding the definition of named entity (NE), we follow [1] who classify NEs into two main categories: generic entities (e.g., person, geographical location and organization) and domain-specific entities (e.g., genes and terms). The increasing need to process a large amount of data facilitates the development of NER models, particularly in specialized fields that exhibit an extensive variety of NEs. Nevertheless, a corpus of high-quality NER annotations is difficult and expensive to obtain since it requires domain expertise. In the practical setting, little or no domain supervision is available, specifically for the Russian language.

Hence, we exploit a method to construct a domain-specific labelled dataset for NER without human supervision. The dataset is then used to train a domain-specific NER model for extraction of knowledge graph (KG) entities in the downstream task of question answering (QA). The methodology includes the following subtasks. (1) Construction of a domain-specific entity vocabulary using the Wikipedia category graph and a set of seed categories. (2) Construction of a domain-specific text corpus from the corresponding Wikipedia articles and publicly available resources. (3) Training a general-domain NER model for preliminary text annotation. (4) Implementation of the morphology-based algorithm for automatic text annotation using the entity vocabulary. (5) Assembly of domain-specific NER dataset from annotations obtained with the general-domain NER model and the morphology-based algorithm. (6) Training the final domain-specific NER model.

We consider that the methodology is domain-transferable and can be modified to depend upon the target domain and available data. We conduct experiments in the domain of Russian History which is represented by a rich set of both generic (historical figures, names of battles, wars and cultural properties) and domain-specific entities (historical terms, concepts and phenomena). Besides, we report the results of the NER model on an additional test set which consists of 1,795 manually annotated samples from the Unified State History Exam.

## 2. RELATED WORKS

Domain-specific NER using external knowledge has been a subject of recent research. The main idea is to use publicly available text data to extract features for NER models [2] or combine texts with non-linguistic data such as game states [3]. Current state-of-the-art NER models incorporate external knowledge from BERT-based language models [4]. Nevertheless, the direct application of pre-trained language models in the domain-specific scenario may result in unsatisfactory performance due to shift in word distribution or an insufficient representation of domain knowledge in the pretraining corpora. The problem has been alleviated in the scientific and biomedical domains for English. SciBERT [5] and BioBERT [6] achieved significant improvement over original BERT [7] on a number of downstream tasks within the domains. However, a vast amount of domain texts is required to train domain-specific language models. Another approach involves training NER models using dictionaries from domain-specific KGs [8]. The model achieved strong performance competitive to supervised benchmarks. Still, there is a lack of publicly available domain-specific KGs and domain-specific NER datasets for the Russian language that can be used as a source of domain knowledge.

A common NER solution is to fine-tune a BERT-based model on the available supervision. However, the target domain usually differs from the pre-training corpus which may result in the unsatisfactory performance of the model on the downstream task. Recently, unsupervised domain adaptation of language models has shown quality improvement in a number of NLP tasks, including sequence labelling [9]. [10] study unsupervised domain adaptation of BERT in the limited labelled and unlabelled data scenarios. The results report that fine-tuning of the pre-trained language model even on a small amount of domain data (1,000 samples) before training on the downstream task improves performance. In our work, we apply the domain adaptation procedure to compare the performance of the trained BERT-based NER models. We now describe the data collection pipeline, automatic annotation procedure and the model training.

## 3. METHOD

We investigate a setting when no domain-specific supervision and annotation resources are available. We proceed from the assumption that a model can learn from another model, and domain-specific texts can be annotated without human effort using only an entity vocabulary. Section 3.1 describes the construction of the domain-specific entity vocabulary V and domain-specific text corpus D using publicly available resources. We also train a general-domain NER model on an available general-domain NER dataset. We refer to this model as RuBERT-general and use it as a source of external knowledge particularly about generic entities (see Section 3.2.1). Besides, we developed a morphology-based annotation algorithm over V which serves as a source of domain knowledge (see Section 3.2.2). Each text from D was annotated with RuBERT-general and the morphology-based annotation algorithm. We then unified the annotations obtained from the previous steps (see Section 3.2.3). The general-domain NER dataset is further combined with the assembled domain-specific NER. This allows us to train a model that is aware of both domain and general knowledge. Finally, we train the domain-specific NER model on the resulting annotations. We describe the training procedure and the results of the experiments in Section 4. Section 5 presents the discussion of the method and outlines future work. Section 6 highlights the main contribution of this work and draws the conclusion.

## 3.1. Data Collection

Data collection pipeline consists of two stages: construction of a domain-specific entity vocabulary V and construction of a domain-specific text corpus D. If a primary V is not available, it is important to conduct the collection procedure thoroughly for the model to learn relevant domain-specific entities as well as contexts in which they can occur.

In the first stage, we used Wikipedia API to retrieve a list of titles of Wikipedia articles related to the domain of Russian History. We traversed the Wikipedia graph over a set of seed categories, e.g. "History of Russia, by periods", "Battles involving Kievan Rus'", "Wars involving Russia" and etc. For each title in the retrieved list, we parsed a text of the corresponding article, a summary, a list of categories and a list of interlinks in the article. Each interlink and Wikipedia title are considered to be NEs and added to V. We collected statistics on how frequently each NE $\in$ V is referred to in the Wikipedia articles. We also built a frequency vocabulary for the list of categories for each NE. NEs with the interlink frequency of lower than 2 and the category frequency of lower than 3 are discarded from V. Furthermore, we filtered V with a set of seed words for category names such as "USSR", "war", "battle", "culture", "politics" and etc. Figure 1 shows an example of NEs for the Wikipedia article "Treaties of Tilsit", where the interlinks are underlined and coloured in blue. The interlink frequency of the NE "Treaties of Tilsit" is 202. The NE relates to the following categories: "19th-century treaties", "Napoleonic Wars treaties", "1807" and etc.

### Treaties of Tilsit
From Wikipedia, the free encyclopedia

*Not to be confused with Act of Tilsit.*

The **Treaties of Tilsit** were two agreements signed by Napoleon I of France in the town of Tilsit in July 1807 in the aftermath of his victory at Friedland. The first was signed on 7 July, between Emperor Alexander I of Russia and Napoleon I of France, when they met on a raft in the middle of the Neman River. The second was signed with Prussia on 9 July. The treaties were made at the expense of the Prussian king, who had already agreed to a truce on 25 June after the Grande Armée had captured Berlin and pursued him to the easternmost frontier of his realm. In Tilsit, he ceded about half of his pre-war territories.[1][*page needed*][2][*page needed*][3]

Figure 1. A summary of the Wikipedia article "Treaties of Tilsit". Each interlink and its corresponding title are added to the domain-specific entity vocabulary.

We additionally retrieved NEs from publicly available structured resources such as glossaries, educational books and tasks from the Unified State History Exam to enrich **V**. Consider the examples of the NEs in the resulting **V**. Generic NEs include "Christianization of Kievan Rus'", "Vladimir the Great", "Seven Years' War", "Saint Petersburg", "Battle of Borodino" and etc. Domain-specific entities include "Oprichnina", "Streltsy", "Bondhold", "Time of Troubles" and etc. An important note should be made that the quality of the **V** depends on the depth of the Wikipedia graph traversal due to a number of ambiguous titles and a large degree of cross-reference in the articles. We experimented with the traversal depth values of 1 and 2. The first vocabulary consists of nearly 17,000 NEs, while the second one includes 95,000 NEs, mostly redundant ones. The quality of the resulted vocabularies was validated manually. Towards a better quality of the domain NEs, we used the first vocabulary in our experiments.

In the second stage, we constructed a domain-specific corpus **D** which consists of texts from publicly available resources such as books about Russian History (autobiographical fiction, educational books, documentaries and series of lectures), the Wikipedia article summaries, the Unified State History Exam variants and tests on Russian History. The size of the corpus is 5.65M tokens. We used **D** to obtain annotations with the general-domain NER model and the morphology-based algorithm. Besides, we used **D** for domain adaptation of the final NER model. We provide details in the next section.

### 3.2. Automatic Annotation Procedure

#### 3.2.1. General-domain Annotation

RuBERT [11] is a monolingual BERT model for the Russian language which outperforms multilingual BERT over a number of NLP tasks for Russian including NER (http://docs.deeppavlov.ai/en/master/features/models/ner.html). In our experiments, we use RuBERT architecture for the **RuBERT-general** model and for the domain-specific NER model. Note that these are two different models. We obtained **RuBERT-general** by fine-tuning RuBERT model on WikiNER [12]. WikiNER is a general-domain NER dataset which has proved its quality for the NER task and is widely used in multiple languages. The dataset consists of 204,778 samples in the Inside-Outside-Beginning (IOB) scheme with 4 semantic labels: "PER", "LOC", "ORG" and "MISC". The data was randomly split into 163,822 train samples, 20,478 dev samples and 20,478 test samples. We trained the model for 5 epochs with default parameters using Hugging Face NLP-library [13]. Evaluation results of the general-domain NER model on the dev and test sets are presented in Table 1.

Table 1. Evaluation of the RuBERT-general model.

|      | Precision | Recall | F1    |
|------|-----------|--------|-------|
| Dev  | 0.910     | 0.914  | 0.912 |
| Test | 0.913     | 0.916  | 0.914 |

Each text ∈ **D** was annotated with **RuBERT-general**. Figure 2 shows an example of the text "Ivan The Terrible introduced oprichnina. " and its **RuBERT-general** annotation. w of indices ∈ {0, …, 4} refers to a word token. "B" (begin) prefix denotes the first token of an entity mention and "I" (inside) prefix corresponds to the tokens following it.

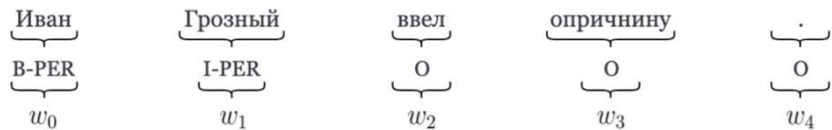

Figure 2. An example of the text annotated with the RuBERT-general model.

#### 3.2.2. Domain-specific Annotation

To provide annotations on domain-specific entities, we implemented a morphology-based algorithm for automatic text annotation using only **V**. The algorithm is closely related to [14]. The work proposes assembly of a general-domain NER dataset, called SESAME, using DBpedia as a source of NEs & their semantic types and Wikipedia as a document collection. A basic idea is to detect mentions of DBpedia entities in a Wikipedia text based on the exact match. Each entity mention (i.e. a character span) is tagged with its corresponding semantic type from DBpedia. Figure 3 shows an example of the annotation. The DBpedia entities "John Smith" and "Rio de Janeiro" are tagged with their corresponding labels "PER" and "LOC" in the IOB format. w of indices ∈ {0, ..., 7} denotes a word token, while s corresponds to a sentence token. In contrast, we construct the entity vocabulary from scratch. Semantic labels are the general-domain NER predictions or the "MISC" label assigned with the morphology-based algorithm. We now describe the annotation procedure with the algorithm.

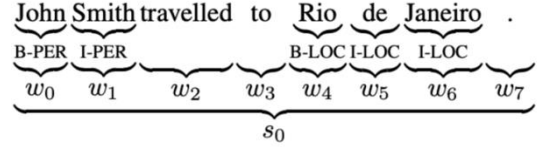

Figure 3. An example of the text annotation algorithm by [14].

In an offline step, we split each text into sentences using rusenttokenize library, a rule-based sentence segmenter for Russian (https://pypi.org/project/rusenttokenize). Each NE ∈ **V** and each text from the corpus was tokenized with Spacy Russian Tokenizer (https://github.com/aatimofeev/spacy_russian_tokenizer) and lemmatized with pymorphy2 [15]. Next, we identify mentions of NEs ∈ **V** in the processed input text. Each mention is tagged with the *"MISC"* label in the IOB format to mark the boundaries of the entity. The remaining tokens are tagged with *"O"* (outside). The overall scheme is outlined in Figure 4, where *t* corresponds to the input text and *e* denotes the detected mentions of entities from **V**.

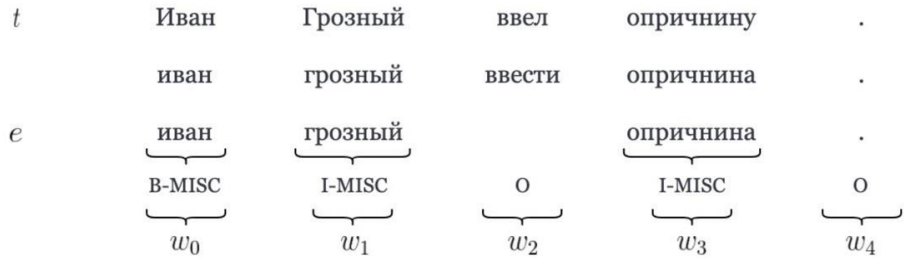

Figure 4. A graphical structure for annotation procedure with the morphology-based algorithm.

Note that one can use a general-domain KG such as Wikidata to map instances of NEs with their corresponding labels to enrich the semantic label inventory, e.g. "city" ⇒ "LOC". Another option for the algorithm modification is to manually annotate all entities in **V** only once, and then assign the labels to all entity mentions. For instance, "Ivan The Terrible" ⇒ "B-PER I-PER" or "Treaties of Tilsit" ⇒ "B-MISC I-MISC I-MISC". During the annotation procedure, the corresponding set of predefined labels is then can be assigned to each entity mention.

### 3.2.3. Annotation Unification

Therefore, each text ∈ **D** has two annotations. The next step is to unify the obtained annotations. If an entity mention is annotated by both the **RuBERT-general** model and the morphology-based algorithm, we prefer the **RuBERT-general** annotation to the morphology-based one. This allows for preserving relevant semantic labels for generic NEs. The unification procedure is illustrated in Figure 5, where the NE "Ivan The Terrible" ($w_0$ and $w_1$) is tagged with the "B-PER" and "I-PER" labels by **RuBERT-general**. The NE "oprichnina" ($w_3$) is tagged with the "MISC" label by the morphology-based algorithm. The remaining tokens are tagged with the "O" label. Annotation obtained with the **RuBERT-general** model is marked as general-domain annotation. Annotation obtained with the morphology-based algorithm is referred to as domain-specific annotation. Annotation unification corresponds to the result of the procedure.

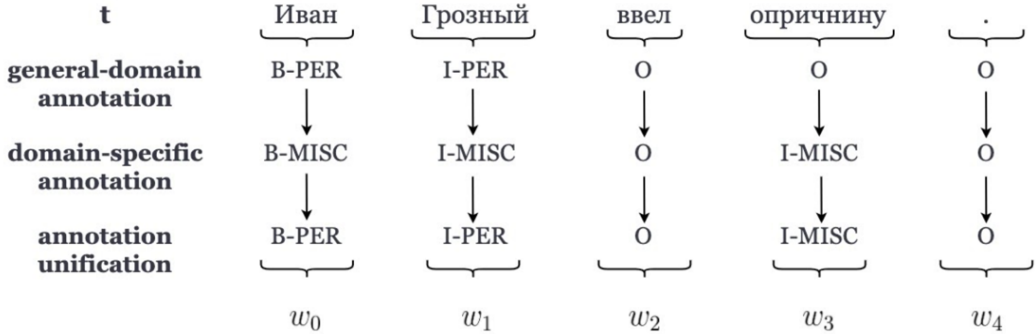

Figure 5. An overall scheme for the annotation unification procedure.

After the unification procedure, we discarded duplicates and samples that are labelled with only the "O" tag. The size of the filtered NER dataset is 160 410 samples. Since the targeted domain exhibits both generic and domain-specific entities, we combined the domain-specific NER dataset with the WikiNER dataset. The total size of the assembled NER dataset is 402,027 samples. Consider an example of the dataset sample "Took part in the Battle of Moscow, the battle of Stalingrad and liberated the Crimea." and its corresponding annotation (see Figure 6).

| Принимал | участие | в | Битве | за | Москву | , |
|---|---|---|---|---|---|---|
| O | O | O | B-MISC | I-MISC | I-MISC | O |
| Сталинградской | битве | и | освобождал | Крым | . | |
| B-MISC | I-MISC | O | O | I-LOC | O | |

Figure 6. A sample from the constructed NER dataset.

## 4. EXPERIMENTS AND RESULTS

In our experiments, we train three BERT-based models on the constructed NER dataset: (1) **BERT-original**, a multilingual BERT model; (2) **RuBERT-original**, a monolingual BERT model for Russian; and (3) **RuBERT-adapted**, an adapted monolingual BERT model for Russian. Specifically, we apply domain adaptation to RuBERT over **D** before training for the NER task. We compare the performance of the three models. To obtain the **RuBERT-adapted** model, we fine-tuned RuBERT language model on **D** using Hugging Face library. The model was trained with default parameters for 12 epochs. The best perplexity achieved is 11.2. Notably, fine-tuning of the language models on domain-specific data may lead to perplexity decrease while increasing the downstream task performance [16]. We trained **BERT-original**, **RuBERT-original** and **RuBERT-adapted** models via Hugging Face library with default parameters for 5 epochs. The training results are shown in Table 2.

Table 2. Evaluation of BERT-original, RuBERT-original and RuBERT-adapted models on the dev and test sets.

| Model | Dev | | | Test | | |
|---|---|---|---|---|---|---|
| | Precision | Recall | F1 | Precision | Recall | F1 |
| BERT-original | 0.908 | 0.908 | 0.908 | 0.907 | 0.906 | 0.907 |
| RuBERT-original | **0.913** | **0.913** | **0.912** | **0.914** | **0.916** | **0.915** |
| RuBERT-adapted | 0.892 | 0.896 | 0.894 | 0.891 | 0.897 | 0.894 |

Knowing that the annotation samples in the collected NER dataset may contain automatic annotation errors or be inconsistent, we manually annotated extra 1,795 samples from the Unified State History Exam. The additional test set allows to assess the performance of the models reliably, disregarding this potential deficiency. The models are evaluated over a full inventory of semantic labels. We computed Precision, Recall and F1-score for each label by taking their average over the weighted number of instances. We present the results for each model in Tables 3, 4 and 5. B and I refer to the beginning and the inside prefixes respectively. AVG is the weighted average metric. Support corresponds to the number of instances.

Table 3. Evaluation of BERT-original on the additional test set over a weighted inventory of semantic labels.

| | B | | | | I | | | | | |
|---|---|---|---|---|---|---|---|---|---|---|
| Metric/Label | LOC | MISC | ORG | PER | LOC | MISC | ORG | PER | O | AVG |
| **Precision** | 0.76 | 0.58 | 0.63 | 0.87 | 0.89 | 0.75 | 0.74 | 0.93 | 0.75 | 0.78 |
| **Recall** | 0.52 | 0.47 | 0.42 | 0.88 | 0.89 | 0.49 | 0.55 | 0.90 | 0.95 | 0.78 |
| **F1** | **0.62** | 0.52 | **0.50** | 0.87 | **0.89** | 0.59 | 0.63 | 0.92 | 0.84 | 0.76 |
| **Support** | 96 | 478 | 105 | 345 | 559 | 1157 | 196 | 617 | 2717 | 6270 |

Table 4. Evaluation of RuBERT-original on the additional test set over a weighted inventory of semantic labels.

| | B | | | | I | | | | | |
|---|---|---|---|---|---|---|---|---|---|---|
| Metric/Label | LOC | MISC | ORG | PER | LOC | MISC | ORG | PER | O | AVG |
| **Precision** | 0.75 | 0.58 | 0.53 | 0.91 | 0.88 | 0.68 | 0.57 | 0.80 | 0.80 | 0.77 |
| **Recall** | 0.49 | 0.49 | 0.33 | 0.71 | 0.87 | 0.51 | 0.56 | 0.94 | 0.94 | 0.78 |
| **F1** | 0.59 | 0.53 | 0.41 | 0.80 | 0.88 | 0.58 | **0.64** | 0.86 | **0.96** | 0.77 |
| **Support** | 96 | 478 | 105 | 345 | 559 | 1157 | 196 | 617 | 2717 | 6270 |

Table 5. Evaluation of RuBERT-adapted on the additional test set over a weighted inventory of semantic labels.

| | B | | | | I | | | | | |
|---|---|---|---|---|---|---|---|---|---|---|
| Metric/Label | LOC | MISC | ORG | PER | LOC | MISC | ORG | PER | O | AVG |
| **Precision** | 0.79 | 0.60 | 0.57 | 0.94 | 0.91 | 0.74 | 0.64 | 0.91 | 0.82 | 0.80 |
| **Recall** | 0.46 | 0.59 | 0.35 | 0.86 | 0.84 | 0.59 | 0.55 | 0.96 | 0.94 | 0.81 |
| **F1** | 0.59 | **0.59** | 0.44 | **0.90** | 0.87 | **0.65** | 0.59 | **0.93** | 0.88 | **0.80** |
| **Support** | 96 | 478 | 105 | 345 | 559 | 1157 | 196 | 617 | 2717 | 6270 |

**BERT-original** demonstrates a slight improvement over **RuBERT-original** model and the best F1-score over "B-LOC", "B-ORG" and "I-LOC" labels. **RuBERT-original** achieves the best F1-score over "I-ORG" and "O" labels. **RuBERT-adapted** shows performance gain over "B-MISC", "I-MISC", "B-PER" and "I-PER" semantic labels which results in +2 Precision score, +3 Recall score and +3 F1-score.

In our work, we aim at extracting mentions of KG entities from texts. In particular, we do not apply semantic labels of NEs, but this can be a useful feature for an entity linking model. Hence, we additionally evaluate the performance over the weighted inventory of 3 semantic labels: "B-MISC", "I-MISC" and "O". The results are shown in Table 6, where Prec. refers to Precision and S corresponds to Support.

Table 6. Evaluation of BERT-original, RuBERT-original and RuBERT-adapted models on the additional test set over a weighted inventory of unified semantic labels.

| Label | **BERT-original** | | | **RuBERT-original** | | | **RuBERT-adapted** | | | S |
|---|---|---|---|---|---|---|---|---|---|---|
| | Prec. | Recall | F1 | Prec. | Recall | F1 | Prec. | Recall | F1 | |
| **B-MISC** | 0.86 | 0.74 | 0.79 | 0.87 | 0.62 | 0.76 | 0.89 | 0.78 | **0.83** | 1024 |
| **I-MISC** | 0.93 | 0.76 | 0.84 | 0.88 | 0.79 | 0.83 | 0.91 | 0.82 | **0.86** | 2529 |
| **O** | 0.77 | 0.95 | 0.85 | 0.80 | 0.94 | 0.86 | 0.82 | 0.94 | **0.88** | 2717 |
| **AVG** | 0.85 | 0.84 | 0.83 | 0.84 | 0.84 | 0.83 | 0.87 | 0.87 | **0.87** | 6270 |

In the unified label setting, **BERT-original** performs on par with **RuBERT-original**. **RuBERT-adapted** achieves performance gains over all semantic labels which results in +3 Precision score, +3 Recall score and + 4 F1-score as compared to **BERT-original** and **RuBERT-original**.

## 5. DISCUSSION

The proposed method to train a domain-specific NER model has received a satisfactory performance with no human effort. The availability of Wikipedia in multiple languages and versatility of the Wikipedia graph may potentially allow for transferability to new domains and across languages. However, a number of drawbacks need to be solved for better performance and generalization of the method:

- The constructed entity vocabulary is not guaranteed to be free of noise. The main reason for irrelevant entities is a large degree of cross-reference in Wikipedia articles. This may cause redundant label predictions in the inference step. The drawback can be alleviated by extracting mentions of entities using a curated list of Wikipedia sections, additional vocabulary filtering or validation by annotators.
- The constructed entity vocabulary suffers from incompleteness. This may be due to the following reasons. First, not all of the interlinks (i.e. NEs) and entity aliases are highlighted in Wikipedia articles. Second, an interlink always refers to one Wikipedia title (i.e. the same surface form of an NE) resulting in a low lexical variability of the entity vocabulary. This can be solved by querying a KG for a set of NE aliases.
- Despite a rich variety of domains covered in Wikipedia, domain-specific knowledge may not be sufficiently represented in the document collection as well as similar publicly available resources. This necessitates the extra data mining for the model to learn relevant contexts.
- Another problem is lemmatization quality. In some cases, incorrect lemmas are obtained due to word ambiguity and the rich inflectional morphology of Russian. Besides, the morphology-based algorithm only partially covers the inventory of semantic labels. Although the majority of such cases are solved during the unification procedure, the remaining part still leads to annotation inconsistency. Hence, the model may get "confused" during the training and inference steps. A solution for this is to postprocess the assembled dataset based on the token-label frequency or use a KG to map ontological types to the corresponding semantic labels.

We believe that the method can be applied in domain-oriented areas, e.g. processing legal documents, educational dialogue systems and QA systems over KGs. Besides, for languages such as English, the method may potentially be transferred with a better performance achieved due to a variety of the available resources that may be used to automatically obtain the supervision.

## 6. CONCLUSIONS

This paper introduces a method to learn a domain-specific NER model for an arbitrary set of named entities without domain-specific supervision available. The code and models used in the experiments can be found at https://github.com/vmkhlv/histqa-domain-ner. The method is based on the semi-supervised approach. Specifically, a document collection can be automatically annotated using natural language pre-processing tools and a domain-specific entity vocabulary which can be constructed from scratch. Besides, we assume that neural models can learn from each other. Pre-trained language models can be used for training a NER model that is aware of both external and domain knowledge. We empirically show that BERT-based models trained over our method receive satisfactory performance with no human effort. However, a number of drawbacks need to be solved to gain performance. Future work is to be dedicated to quality improvement and exploring the transferability of the method across multiple domains and languages. The latter can be obtained thanks to versatility of Wikipedia.